\documentclass[letterpaper,10 pt,conference]{ieeeconf}
\IEEEoverridecommandlockouts
\let\labelindent\relax
\usepackage{graphicx} \graphicspath{ {figures/} }
\usepackage{amsmath,amssymb,mathabx}
\usepackage{algorithmic}
\usepackage[ruled]{algorithm2e}
\usepackage{acronym}
\usepackage{enumitem}
\usepackage{booktabs}
\usepackage{stfloats}
\usepackage{hyperref}
\usepackage{balance}
\usepackage{xspace,setspace}
\usepackage[skip=3pt,font=small]{subcaption}
\usepackage[skip=3pt,font=small]{caption}
\usepackage[dvipsnames]{xcolor}
\usepackage[capitalise]{cleveref}
\usepackage{tabularx,colortbl,multirow,array,makecell}
\usepackage{overpic}
\usepackage{cite}
\usepackage{tikz}
\usepackage[]{mdframed}
\usepackage{soul}
\usepackage{anyfontsize}

\newcommand*\circled[1]{\tikz[baseline=(char.base)]{
            \node[shape=circle,draw,inner sep=0.6pt] (char) {#1};}}

\makeatletter
\DeclareRobustCommand\onedot{\futurelet\@let@token\@onedot}
\def\@onedot{\ifx\@let@token.\else.\null\fi\xspace}
\def\eg{\emph{e.g}\onedot}

\def\etal{\emph{et al}\onedot}
\makeatother

\crefname{algocf}{alg.}{algs.}
\Crefname{algocf}{Algorithm}{Algorithms}

\makeatletter
\def\BState{\State\hskip-\ALG@thistlm}
\makeatother

\makeatletter
\renewcommand{\paragraph}{%
  \@startsection{paragraph}{4}%
  {\z@}{0ex \@plus 0ex \@minus 0ex}{-1em}%
  {\hskip\parindent\normalfont\normalsize\bfseries}%
}
\makeatother

\crefname{algocf}{alg.}{algs.}
\Crefname{algocf}{Algorithm}{Algorithms}

\definecolor{gblue}{HTML}{4285F4}
\definecolor{gred}{HTML}{DB4437}

\acrodef{dof}[DoF]{Degree of Freedom}
\acrodef{vkc}[VKC]{Virtual Kinematic Chain}
\acrodef{tamp}[TAMP]{Task and Motion Planning}
\acrodef{pddl}[PDDL]{Planning Domain Definition Language}
\acrodef{rrt}[RRT]{Rapidly-exploring Random Tree}
\acrodef{ompl}[OMPL]{Open Motion Planning Library}
\acrodef{iws}[IWS]{Iterated Width Search}
\acrodef{bfs}[BFS]{Breadth First Search}
\acrodef{ai}[AI]{Artificial Intelligence}
\acrodef{vln}[VLN]{Vision-Language Navigation}
\acrodef{3dsg}[3DSG]{graph-based scene representation}
\acrodef{llm}[LLM]{Large Language Model}
\acrodef{pog}[PoG]{Planning on Graph}
\acrodef{epog}[EPoG]{Exploration and Planning on Graph}
\acrodef{ged}[GED]{Graph Edit Distance}

\newcommand{\epog}{EPoG\xspace}

\newcommand{\framedtext}[1]{%
\par\vspace{1mm} 
\noindent\fbox{%
    \parbox{\dimexpr\linewidth-2\fboxsep-2\fboxrule}{#1}%
}%
\par\vspace{2mm} 
}

\definecolor{gblue}{HTML}{4285F4}
\definecolor{gred}{HTML}{DB4437}

\definecolor{custorange}{RGB}{255, 147, 30}
\definecolor{custblue}{RGB}{63, 167, 243}
\definecolor{custdarkblue}{RGB}{38, 99, 145}
\definecolor{custgrey}{RGB}{202, 202, 202}
\definecolor{custgreen}{RGB}{34, 139, 34}

\colorlet{lightorange}{custorange!20}
\newcommand{\hllo}[1]{%
    {%
    \sethlcolor{lightorange}%
    \hl{#1}%
    }%
}
\colorlet{lightblue}{custblue!20}
\newcommand{\hllb}[1]{%
    {%
    \sethlcolor{lightblue}%
    \hl{#1}%
    }%
}
\colorlet{lightgrey}{custgrey!40}

\title{\LARGE \bf Integrated Exploration and Sequential Manipulation on Scene Graph with LLM-based Situated Replanning}

\author{Heqing Yang$^{1,2}$\quad{}Ziyuan Jiao$^{1,2\dagger}$\quad{}Shu Wang$^{3}$\quad{}Yida Niu$^{2,4}$\quad{}Si Liu$^{1\dagger}$\quad{}Hangxin Liu$^{2}$
\thanks{$^\dagger$~Corresponding authors. This work was conducted during Heqing Yang's internship at the Beijing Institute for General Artificial Intelligence (BIGAI). $^{1}$~Beihang University. $^{2}$~State Key Laboratory of General Artificial Intelligence, BIGAI, Beijing, China. $^{3}$~University of California, Los Angeles. $^{4}$~Institute for Artificial Intelligence, Peking University.}%
}
\begin{document}

\maketitle
\thispagestyle{empty}
\pagestyle{empty}

\begin{abstract}
In partially known environments, robots must combine exploration to gather information with task planning for efficient execution. 
To address this challenge, we propose \epog, an \underline{E}xploration-based sequential manipulation \underline{P}lanning framework \underline{o}n Scene \underline{G}raphs. 
\epog integrates a graph-based global planner with a \ac{llm}-based situated local planner, continuously updating a belief graph using observations and \ac{llm} predictions to represent known and unknown objects. 
Action sequences are generated by computing graph edit operations between the goal and belief graphs, ordered by temporal dependencies and movement costs. This approach seamlessly combines exploration and sequential manipulation planning. 
In ablation studies across 46 realistic household scenes and 5 long-horizon daily object transportation tasks, \epog achieved a success rate of 91.3\%, reducing travel distance by 36.1\% on average. 
Furthermore, a physical mobile manipulator successfully executed complex tasks in unknown and dynamic environments, demonstrating \epog's potential for real-world applications.
\end{abstract}

\setstretch{0.94}

\section{Introduction}
To autonomously perform complex tasks, robots require an environment representation that is both expressive and readily queryable for planning. Recently, graph-based scene representations have emerged as a unifying paradigm~\cite{armeni20193d}, modeling scenes from 3D perception~\cite{han2021reconstructing,rosinol2021kimera,wu2023incremental} and augmenting them with predicate-like attributes~\cite{han2022scene,agia2022taskography,rosinol20203d}. These representations facilitate action-centric reasoning~\cite{rana2023sayplan} by grounding symbolic structure in perception, reducing the burden of manual domain engineering, and supporting complex manipulation tasks that are difficult to specify or solve with traditional pipelines~\cite{chen2022think,jiao2022sequential,gu2023conceptgraphs,liu2023bird}.

Existing planning methods that leverage graph-based scene representations largely target two classes of problems: exploration and sequential manipulation. In exploration, robots navigate unknown or partially observed environments for objectives such as mapping~\cite{rosinol2021kimera} or object search~\cite{gu2023conceptgraphs}. These settings typically involve limited physical interaction with the scene, which constrains the range of tasks that can be executed. Sequential manipulation methods, in contrast, explicitly model interaction~\cite{jiao2022sequential,agia2022taskography} but often assume a fully known, static environment.

\begin{figure}[!t]
    \centering
    \includegraphics[width=\linewidth]{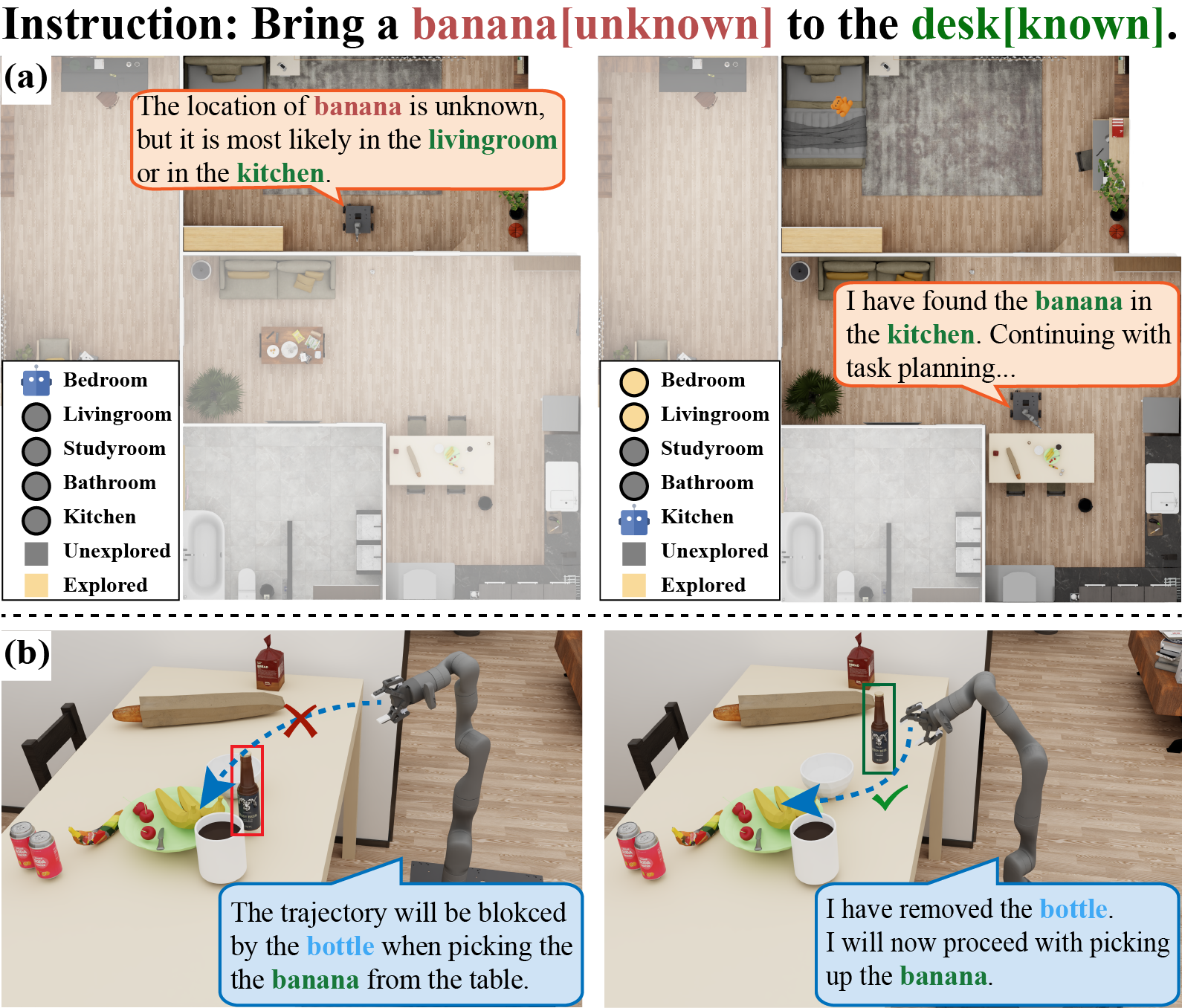}
    \caption{\textbf{An example illustrating the challenges of integrating exploration and sequential manipulation:} (a) Robots must prioritize potential exploration locations and balance exploration with manipulation tasks to execute efficiently. (b) The robot needs to engage in situated planning to handle unexpected situations.}
    \label{fig:motivation}
    \vspace{-24pt}
\end{figure}

Real-world deployment violates these assumptions. A robot may have incomplete prior knowledge and may not immediately observe changes induced by human activity~\cite{hopko2022human}. Consequently, it must interleave information gathering with task planning, which introduces three coupled challenges: 1)~locating task-relevant objects under partial observability, requiring prioritized exploration; 2)~trading off exploration and manipulation to reduce overall execution cost; and 3)~reasoning under uncertainty to produce situated, executable plans. \cref{fig:motivation} illustrates these challenges. Although prior work addresses subsets of them~\cite{yang2024text2reaction,do2023information,garrett2020online}, existing approaches often depend on hand-engineered heuristics, which limit robustness and scalability in long-horizon tasks.

\begin{figure*}[t!]
\centering
\includegraphics[width=0.96\linewidth]{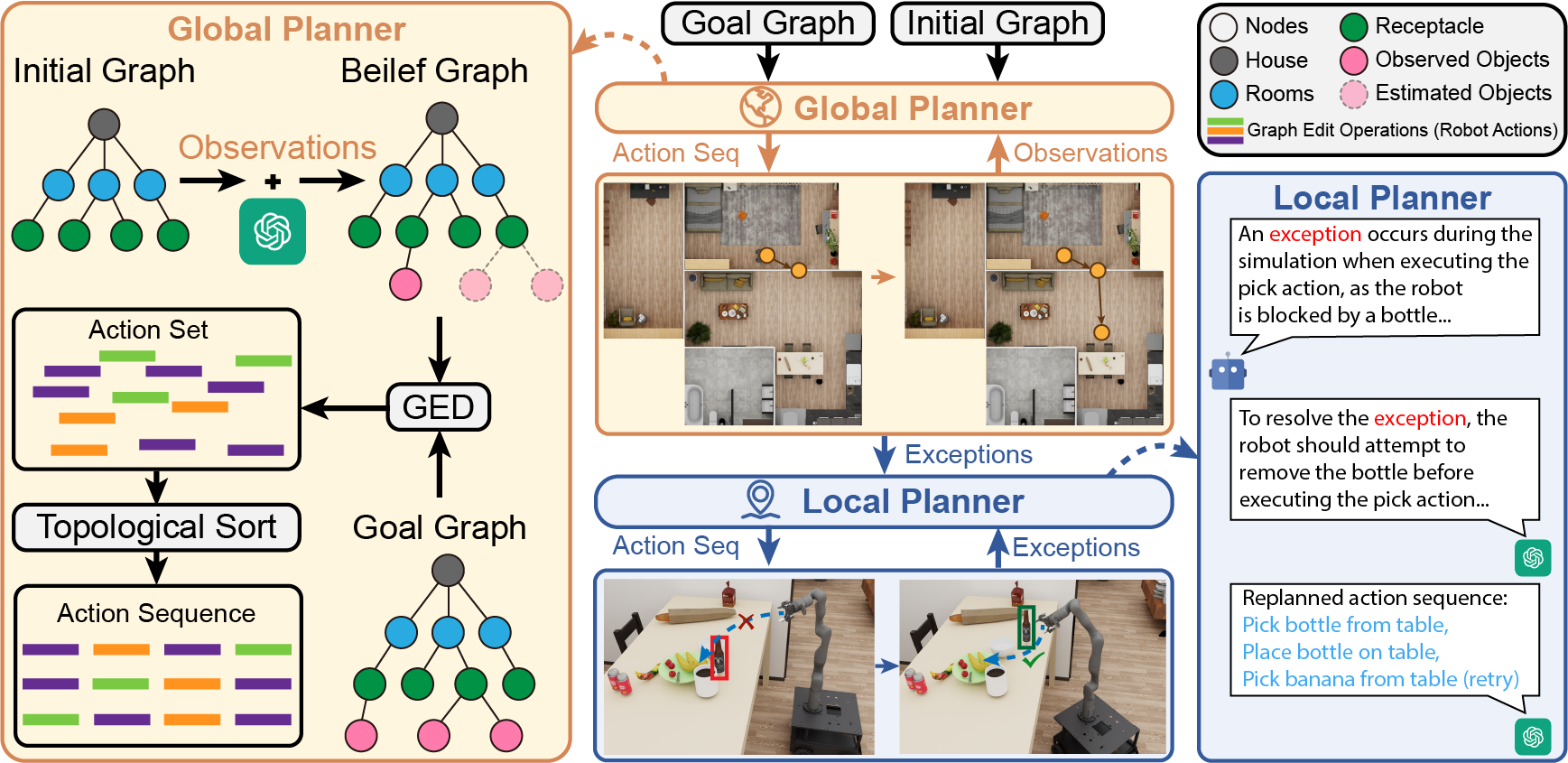}
\caption{\textbf{Overview of the proposed \epog framework.} In the global planner, the belief graph is updated by estimating the locations of target objects present in the goal graph but missing in the initial graph, using new observations and \ac{llm} predictions. The action plan is then obtained by performing the topological sort on the graph edit operations between the belief and the goal graph. In the local planner, the \ac{llm} generates a situated action sequence to handle exceptions encountered during execution.}
\label{fig:framework}
\vspace{-20pt}
\end{figure*}

To address these challenges, we propose \epog, a framework that integrates \underline{E}xploration and sequential manipulation \underline{P}lanning \underline{o}n Scene \underline{G}raphs. \epog adopts a bilevel planning architecture for partially observed environments, leveraging pretrained \acp{llm} for informed exploration and situated replanning. As shown in \cref{fig:framework}, the global planner incrementally constructs a belief graph from onboard observations and \ac{llm}-based predictions, computes graph edit operations between the belief and goal graphs, and generates a candidate action sequence via topological sorting. The robot executes this sequence while continuously updating the belief graph with new observations and \ac{llm} predictions, thereby interleaving exploration and manipulation in a closed loop. When execution deviates from the nominal plan, the local planner invokes \acp{llm} for situated replanning to resolve exceptions.
We evaluate \epog on five long-horizon object-transport tasks across 46 household scenes from the ProcThor-10k dataset~\cite{deitke2022}. Quantitative results demonstrate the effectiveness of \epog, and ablations show that: i)~the proposed formulation naturally couples exploration and manipulation, reducing total execution effort; ii)~\epog outperforms purely \ac{llm}-based planners on long-horizon tasks, achieving substantially higher success rates; and iii)~\ac{llm}-guided heuristics and the low-level planner reduce manual design effort while improving execution efficiency. Experiments on a mobile robot further validate real-world applicability.

\subsection{Related Work}
\textbf{Task planning with \acp{llm}} is increasingly popular in robotics for interpreting natural-language instructions and leveraging broad commonsense priors~\cite{achiam2023gpt,zhao2024large,wang2024llm}. They can translate high-level commands into executable action sequences~\cite{song2023llm}, enabling flexible task specifications across diverse environments. However, \acp{llm} remain brittle on long-horizon problems due to limited spatial grounding, imperfect handling of temporal dependencies, and difficulty reasoning under environmental uncertainty~\cite{kambhampati2024llms,valmeekam2022large,yamadaevaluating}. To mitigate these limitations, recent work couples \acp{llm} with classical planning. For example, Zhou~\etal~\cite{zhou2024isr} use \acp{llm} to generate \ac{pddl} domain descriptions, while other approaches use \acp{llm} as heuristics to guide search or exploration~\cite{zhao2024large,shah2023navigation}. Such hybrid systems combine the contextual flexibility of \acp{llm} with the reliability and verifiability of symbolic planners. Nevertheless, most of these methods operate over predicate-based symbolic states, leaving task planning over graph-structured scene representations comparatively underexplored. Graph-based representations~\cite{rosinol2021kimera,chen2022think,jiao2022sequential,agia2022taskography,gu2023conceptgraphs,liu2023bird} offer a complementary substrate by integrating perception with structured spatial and semantic relations, often providing richer geometric context than flat predicates. This motivates studying how \acp{llm} can be integrated with graph-based planning: \acp{llm} can supply context-dependent priors and action proposals, while graph structure provides grounded relational context for reasoning, constraint checking, and replanning. Together, these ingredients may yield more robust and efficient planning in complex, partially observed environments.

\textbf{Task planning using scene graphs} primarily falls into two distinct categories: methods based on known scene graphs that leverage predefined information for task planning~\cite{gu2023conceptgraphs,ni2024grid}; and methods based on unknown scene graphs, which involve robot exploration and the creation of new graphs for dynamic task execution~\cite{rana2023sayplan}. However, these approaches face three key limitations: 1)~Current methodologies predominantly utilize 2D topological relationships while neglecting essential 3D spatial attributes (\eg, volumetric occupancy and geometric constraints), which can compromise task feasibility; 2)~the inherent context length limits of transformer-based models, such as \acp{llm}, impose restrictions on processing large-scale graphs, which is common for real-world environments; 3)~current planning schemes for graph-based scene representations inadequately capture spatial-temporal relationships between scene entities that evolve over time or by human activities. These challenges motivate the investigation of new frameworks that can effectively harness both geometric and semantic information from graph-based scene representations while addressing computational scalability requirements.

\textbf{Task planning in unknown environments} requires robots to dynamically adapt plans based on new observations and continuously update their understanding of the environment~\cite{rintanen2004complexity,rosinol2021kimera,yao2022react}. Effective planning in such settings necessitates integrating exploration with task execution, balancing the acquisition of missing information with goal achievement. Existing approaches either explore the entire environment before planning~\cite{jiang2024roboexp} or rely on manually defined strategies~\cite{xu2022framework}, resulting in inefficiency and limited adaptability. In contrast, our work integrates exploration and planning using a graph-based representation that accounts for movement costs and employs \acp{llm} for local replanning to handle exceptions, improving execution efficiency and versatility in dynamic environments.

\setstretch{0.95}

\section{Preliminary} \label{sec:prelim}
In this section, we first introduce the graph-based representation used in this work and then elaborate on how the planning problem is formulated using this representation.

\subsection{Graph-Based Scene Representation}
Following~\cite{han2021reconstructing,han2022scene,jiao2022sequential}, the graph-based scene representation is structured as a tree, providing a hierarchical semantic abstraction of the environment. A scene graph $G=(V, E, A)$ consists of scene nodes $V$, edges $E$, and task-dependent attributes $A$ associated with the nodes, which define the potential interactions with each node.

\textbf{Scene Nodes} $v_i\in V$ represent the entities in the scene, where $v_i=\langle o_i, c_i, M_i, B_i \rangle$. Here, $o_i$ is the object's unique identifier in the graph, $c_i$ is its semantic label, $M_i = \{m_i^j \mid j = 1, \ldots, |M_i|\}$ is the set of geometric primitives describing its full geometry, and $B_i$ is its 3D bounding box.

\textbf{Scene Edges} $e_{i,j} \in E$ are directed edges representing the supporting relationship between nodes. Each edge $e_{i,j} = \langle v_i, v_j, t_{i,j}, c_{i,j} \rangle$ is defined by a spatial transformation $t_{i,j}$ from the parent node $v_i$ to the child node $v_j$ and a semantic description $c_{i,j}$ of the supporting relationship, such as $c_{i,j} \in \{\texttt{on, contain}\}$.

\textbf{Attributes} $A = \{(A_i^s, A_i^c) \mid i \leq |V|\}$ represent the attributes assigned to each node $v_i$, where $A_i^s$ is a supporting attribute indicating whether an object $v_i$ can support another. The attribute $A_i^c$ is optional and only assigned to container objects. Both attributes are used to verify whether a supporting relationship $c_{i,j}$ holds, which helps determine the feasibility of an action during task planning.

\begin{figure}[t!]
\centering
\includegraphics[width=\linewidth]{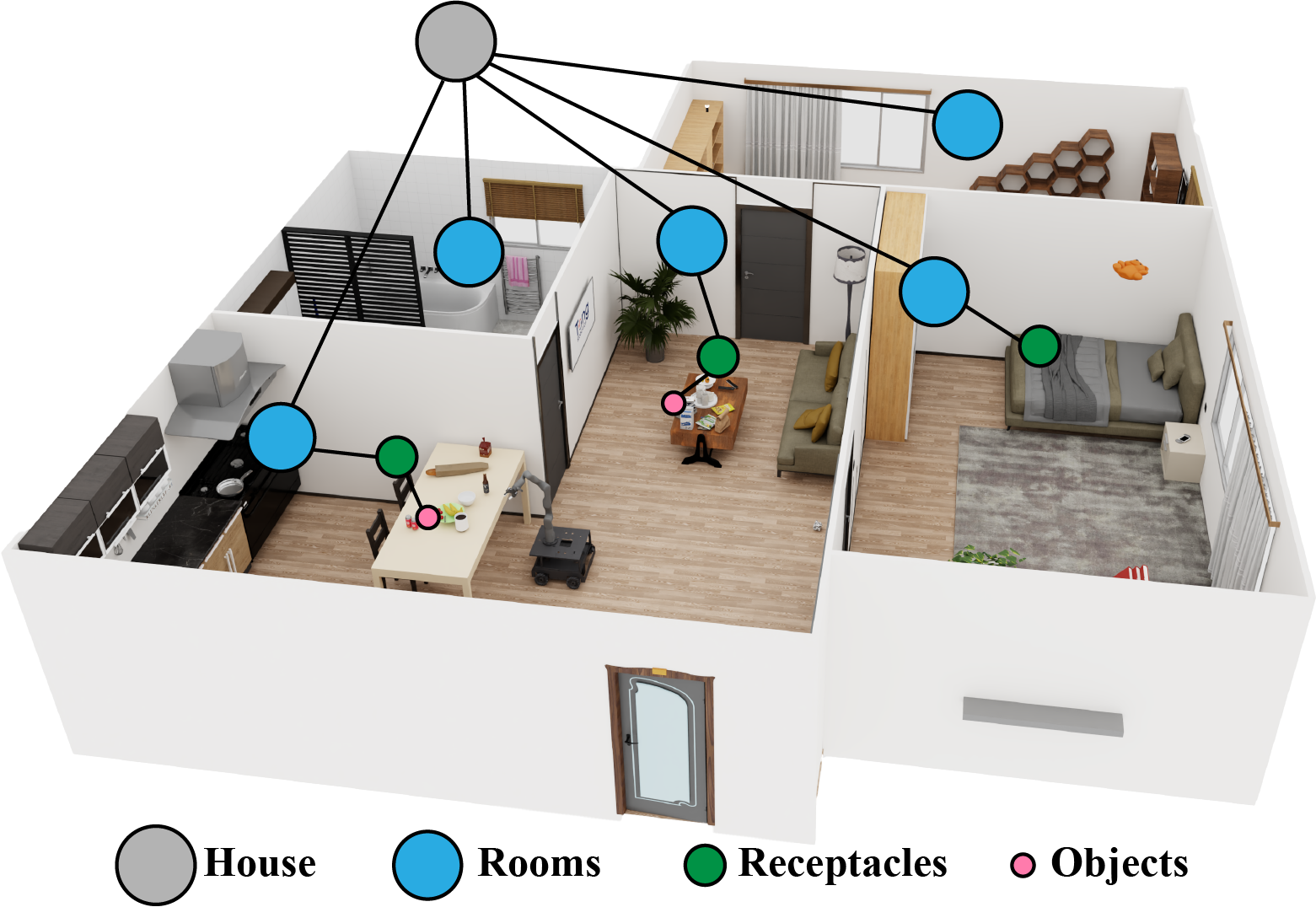}
\caption{An example illustration of an indoor scene.}
\label{fig:scene_graph}
\vspace{-16pt}
\end{figure}

Based on the above definition, in this paper, an indoor scene is hierarchically structured into four levels: \textit{Level 0:} a \texttt{House} as the root; \textit{Level 1:} \texttt{Rooms} within the \texttt{House}; \textit{Level 2:} \texttt{Receptacles} like containers and surfaces within the rooms, which serve as support structures; and \textit{Level 3+:} \texttt{Objects}, which are the primary task targets. This hierarchical mapping supports downstream tasks and has demonstrated good results. \cref{fig:scene_graph} provides an example illustration of an indoor scene with this structure.

\subsection{Problem Definition}

In this work, we address the problem of sequential manipulation planning in a setting with unknown object states, a finite number of actions, and deterministic transitions. This problem $\mathcal{P}$ is represented by a tuple $\langle \mathcal{S}, \mathcal{A}, \mathcal{T}, \mathcal{O}, s_{init}, s_{goal} \rangle$. Given an environment state \( s \in \mathcal{S} \), an action \( a \in \mathcal{A} \) can be selected from the set of applicable actions. The transition function \( T: s \xrightarrow{a_i} s' \) defines the dynamics of the environment, indicating that executing action \( a_i \) in the state \( s \) results in a new state \( s' \). The details of these actions will be introduced in \cref{sec:method}. The observation $obs=\{V_{obs},E_{obs}\}\in\mathcal{O}$ is what the robot perceives upon reaching a new state. A solution to the planning problem $\mathcal{P}$ is a sequence of actions $\pi = (a_1, \ldots, a_k)$ that transforms the initial state $s_{init}$ to the goal state $s_{goal}$. 

In this paper, the state $s$ is defined as $G$. The initial state $s_{init}$ typically includes only \texttt{House}, \texttt{Room}, and \texttt{Receptacle} nodes, as well as the edges between them, excluding \texttt{Object} nodes, assuming the robot does not know object locations. This assumption is valid, as rooms and large objects are less likely to be frequently moved by humans, whereas smaller objects may be. The goal state $s_{goal}$ is also represented as a graph, including all task-relevant nodes and their relationships. Object relationships are simplified for clarity into semantic descriptions, \eg, \texttt{apple} on \texttt{table}.

\section{\epog Framework} \label{sec:method}

The proposed \epog framework addresses task planning for robots in partially known environments using scene graphs as the planning representation. It integrates exploration and sequential manipulation to generate efficient action sequences, enabling robots to explore unknown environments while completing tasks. A pre-trained \ac{llm} facilitates situated replanning to handle unexpected situations, improving robustness and reducing design effort. The following sections detail the framework and its components.

\subsection{\epog Overview}
\cref{alg:epog} outlines the \epog framework, which utilizes a bi-level planning scheme. The global planner begins with an initial belief graph representing the environment, combining known and unknown objects and their relationships. This graph is built from prior knowledge, including object locations, relationships, and spatial constraints, while accounting for uncertainty in areas that the robot has yet to explore, using commonsense knowledge from \acp{llm}. As the robot executes tasks and gathers new observations, the belief graph is dynamically updated, supporting long-horizon planning through \ac{ged} and topological sorting. The process iterates until task completion. If exceptions occur during execution, an \ac{llm}-based local planner recursively resolves the issues until the action succeeds. The following sections detail the global and local planners.

\SetKwFunction{LocalPlanner}{LocalPlanner}
\SetKwFunction{GraphBasedPlanner}{GraphBasedPlanner}
\SetKwFunction{UpdateGraph}{UpdateGraph}
\SetKwFunction{LLMPlanner}{LLMPlanner}
\SetKwFunction{EstimateBeliefGraph}{EstimateBeliefGraph}
\SetKwFunction{Pop}{Pop}
\SetKwFunction{RollOut}{RollOut}

\subsection{Global Planner} \label{sec:global_planner}
\EstimateBeliefGraph{$\cdot$} estimates the distribution of task-relevant objects and their relationships before planning. Initially, the belief graph may miss nodes and edges for task-relevant objects. For each missing node, a pre-trained \ac{llm} infers probable object locations, which are added to the belief graph. Two prompts are generated per object: (1) estimate the \texttt{Room} where the object is likely located, and (2) predict the \texttt{Receptacle} where it might be placed. To complete the belief graph $G_b$, we add task-relevant nodes $V_{task}$ and attributes $A_{task}$ from $G_g$, and edges $E_{est}$ based on the predicted relationships. For each edge \( e_i = \langle v_i, v_j, t_{i,j}, c_{i,j} \rangle \), we do not predict the transformation \( t_{i,j} \), as the object pose remains unknown until detection. In addition, we assume that the information for missing nodes \( v_i = \langle o_i, c_i, M_i, B_i \rangle \) is given in the $G_g$. The final description of the updated belief graph is thus \( G_b = (V_{init} \cup V_{task}, E_{init} \cup E_{est}, A_{init} \cup A_{task}) \).

\RollOut{$\cdot$} is performed using a motion planner to generate the motion plan~\cite{wang2024llm} before executing an action. If the motion planner raises an exception, the action will not be executed. Otherwise, it is carried out in the environment. After execution, the robot receives a new observation \( obs = \{V_{obs}, E_{obs}\} \), where \( V_{obs} \) represents the visible objects and \( E_{obs} \) denotes the relationships between them. In this work, we assume that whenever the robot enters a \texttt{Room}, it can observe all objects and their relationships, except those contained within a closed \texttt{Receptacle}.

\begin{algorithm}[t!]
    \SetKwInOut{Input}{Input}\SetKwInOut{Output}{Output}
    \SetKwFunction{Add}{Add}
    \SetKwProg{Fn}{Function}{:}{}
    \caption{\epog}
    \label{alg:epog}
    \LinesNumbered
    \small
    \Input{Initial Graph $G_{init}$; Goal Graph $G_{g}$}
    \textcolor{blue}{// Global Planner}\\
    \textcolor{blue}{// Initial estimation of the belief graph}\\
    $G_{b}$ $\leftarrow$ \EstimateBeliefGraph{$G_{init}$}\;
    \textcolor{blue}{// Planning on scene graph; see \cref{alg:optimized_planner}}\\
    $P_{g}$ $\leftarrow$ \GraphBasedPlanner{$G_{b}$, $G_{g}$}\;
    \While{$P_{g}$ is not empty}{
        $action$ $\leftarrow$ \Pop{$P_{g}$}\;
        $obs, exception$ $\leftarrow$ \RollOut{$action$}\;
        $G_{b}$, $replan$ $\leftarrow$ \UpdateGraph{$G_{b}$, $obs$}\;
        \If{$replan$ is not $none$}{
            $P_{g}$ $\leftarrow$ \GraphBasedPlanner{$G_{b}$, $G_{g}$}\;
        }\ElseIf{$exception$ is not $none$}{
            \LocalPlanner{$exception$}\;
        }
    }
    \textcolor{blue}{// Local Planner}\\
    \Fn{\LocalPlanner{$exception$}}{
        \textcolor{blue}{// \ac{llm}-based planner; see \cref{sec:local_planner}}\\
        $actions$ $\leftarrow$ \LLMPlanner{$exception$}\;
        \ForEach{$action$ \textbf{in} $actions$}{
            $obs, exception$ $\leftarrow$ \RollOut{$action$}\;
            \textcolor{blue}{// Recursively resolving exceptions}\\
            \If{$exception$ is not $none$}{
                \LocalPlanner{$exception$}\;
            }
        }
    }
\end{algorithm}

\UpdateGraph{$\cdot$} maintains the belief graph in sync with current observations, $\mathcal{O} = (V_{obs}, E_{obs})$, where $V_{obs}$ represents visible objects (nodes) and $E_{obs}$ captures their spatial and semantic relationships (edges). If a target object's initial position estimate $e_\text{err}$ is absent from $V_{obs}$, the global planner updates the belief graph by replacing $(v_i, e_{err})$ with a new estimate $(v_i, e_{new})$ based on $G_b$. For instance, if an apple is expected on a table but is not found there, the planner re-estimates its location.

\begin{algorithm}[t!]
\SetKwFunction{ComputeGED}{GED}
\SetKwFunction{GetConstraints}{GetConstraints}
\SetKwFunction{InsertWalk}{InsertWalkActions}
\SetKwFunction{Heuristic}{HeuristicCost}
\SetKwFunction{GoalCheck}{IsGoalReached}
\SetKwFunction{NextActions}{NextActions}
\SetKwFunction{UnconstrainedActions}{UnconstrainedActions}
\SetKwProg{Function}{Function}{:}{}
\SetKwInOut{Input}{Input}\SetKwInOut{Output}{Output}
\SetKw{Break}{break}
\SetKw{Continue}{continue}
\LinesNumbered
\small
\Input{Belief graph $G_b$, Goal graph $G_g$}
\Output{Optimal plan $\pi^*$}
\caption{Optimized Graph-Based Task Planner}
\label{alg:optimized_planner}

\textcolor{blue}{// Step 1: Generate primitive actions} \\
$K \leftarrow$ \ComputeGED{$G_b$, $G_g$} \tcp*{min-cost edit operations}
$C \leftarrow$ \GetConstraints{$K$} \tcp*{constraints between actions}

\textcolor{blue}{// Step 2: Initialize search frontier} \\
Stack $\mathit{stack} \leftarrow \emptyset$ \\
$\mathit{min\_cost} \leftarrow \infty$ \\
$\pi^* \leftarrow \emptyset$ \\
\ForEach{$a \in$ \UnconstrainedActions{$(K,C)$}}{
    $\mathit{stack}$.push(([a], $K \setminus \{a\}$, $C \setminus \{a\}$)) \tcp*{Search Node}
}

\textcolor{blue}{// Step 3: Depth-first search} \\
\While{$\neg\mathit{stack}.empty()$}{
    ($\pi$, $K_r$, $C_r$) $\leftarrow \mathit{stack}$.pop() \\
    $\mathit{cost} \leftarrow $\Heuristic{$\pi$}  \\
    \textcolor{blue}{// Prune suboptimal branches} \\
    \eIf{$\mathit{cost} \geq \mathit{min\_cost}$}{
        \Continue
    }{
        \textcolor{blue}{// Goal check} \\
        \If{\GoalCheck{$\pi$}}{
            $\pi' \leftarrow$ \InsertWalk{$\pi$} \\
            \If{$\mathit{cost} < \mathit{min\_cost}$}{
                $\mathit{min\_cost} \leftarrow \mathit{cost}$ \\
                $\pi^* \leftarrow \pi'$
            }
            \Continue
        }
        \textcolor{blue}{// Expand successors} \\
        \ForEach{$a \in$ \NextActions{$K_r$, $C_r$}}{
            $\pi_{\mathit{new}} \leftarrow \pi \oplus a$ \tcp*{append action}
            $K_{\mathit{new}} \leftarrow K_r \setminus \{a\}$ \tcp*{remove action}
            $C_{\mathit{new}} \leftarrow C_r \setminus \{a\}$ \tcp*{remove constraints}
            $\mathit{stack}$.push(($\pi_{\mathit{new}}$, $K_{\mathit{new}}$, $C_{\mathit{new}}$)) 
        }
    }
}
\Return{$\pi^*$}
\end{algorithm}

\GraphBasedPlanner{$\cdot$}, as detailed in \cref{alg:optimized_planner}, is invoked after updating the belief graph to generate a task plan with topological sort. Given an initial and goal graph, \ac{ged} provides a set of graph edit operations that transform the initial graph into the goal graph with minimal action cost. Formally,
\begin{equation}
    GED(G_1, G_2)=\min_{\{a_1, \ldots, a_k\} \in \mathcal{K}\left(G_1, G_2\right)} \sum_{i=1}^k cost\left(a_i\right)
\end{equation}
where $\{a_1, \ldots, a_k\} \in \mathcal{K}\left(G_1, G_2\right)$ denotes a set of edit operations transforming $G_1$ into $G_2$, and $cost(a) \geq 0$ is the cost of each graph edit operation $a$. We consider four types of edit operations, corresponding to robot actions:

\begin{itemize}[leftmargin=*,noitemsep,nolistsep,topsep=0pt]
    \item delete($e_{i,j}$) $\rightarrow$ \texttt{Pick}($v_i$, $v_j$): Pick object $v_j$ from $v_i$.
    \item insert($e_{i,j}$) $\rightarrow$ \texttt{Place}($v_i$, $v_j$): Place object $v_j$ on $v_i$.
    \item substitute($A_i^c$, \texttt{opened}) $\rightarrow$ \texttt{Open}($v_i$): Open door $v_i$ to make contained objects accessible and observable.
    \item substitute($A_i^c$, \texttt{closed}) $\rightarrow$ \texttt{Close}($v_i$): Close door $v_i$ to make contained objects inaccessible and unobservable.
\end{itemize}

Based on this correspondence, we use \ac{ged} to obtain a set of necessary graph edit operations between the belief and goal graphs, resulting in a set of unsorted actions \( K \). These actions have temporal dependencies, forming a partially ordered set. For example, for node \( v_i \), the action \texttt{pick}(\(v_i\), \(v_j\)) must occur before \texttt{place}(\(v_i\), \(\cdot\)). These temporal dependencies create a constraint set \( C = \{(a_i, a_j) \mid i \neq j \} \), where each action pair \( (a_i, a_j) \) must satisfy the condition \( a_i < a_j \), meaning that $a_i$ must occur before $a_j$. This ensures that actions are performed in the correct sequence, respecting the temporal order necessary to complete the task. 

The task planning problem on $G$ can then be formulated as a topological sorting problem on $(K, C)$, to generate a sorted sequence of actions $\pi$. To address this, we define a search node $\mathcal{N} = \left(K^{\prime}, C^{\prime}, \pi^{\prime}\right)$, where $K^{\prime} \subseteq K$ represents the unexplored actions, $C^{\prime} \subseteq C$ denotes the remaining temporal constraints, $\pi^{\prime} \subseteq \pi$ is the subsequence of the action sequence. $K \setminus \{a\}$ is the action set except for the action $\{a\}$.

\Heuristic($\cdot$) estimates the robot's travel distance for the current action sequence. Since some object poses are unknown to the robot, a \texttt{Walk} action is inserted where the robot first moves near its parent \texttt{Receptacle} node. The simple yet effective $A^\star$ algorithm is used to estimate the travel distance between actions. As a result, \GraphBasedPlanner accounts for the robot's movement cost and seeks to minimize travel distance. Additionally, a pruning strategy is employed to continuously update the upper bound of the moving cost, effectively reducing the search space and optimizing the path planning.

\begin{figure}[t!]
\centering
\includegraphics[width=\linewidth]{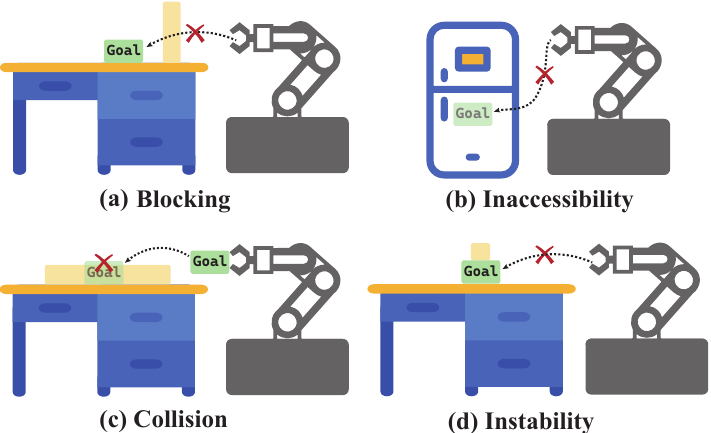}
\caption{\textbf{Examples of exceptions in motion planning.} (a) \textbf{Blocking}: The robot must avoid collisions with other objects while placing or picking up objects. (b) \textbf{Inaccessibility}: Successful object retrieval or placement within a container requires the container to be opened first. (c) \textbf{Collision}: The robot must ensure that placed objects do not collide with the environment. (d) \textbf{Instability}: The robot must maintain the stability of stacked objects during manipulation; for example, retrieving a book beneath a cup may cause instability.}
\label{fig:exception}
\vspace{-16pt}
\end{figure} 

\subsection{Local Replanning} \label{sec:local_planner}
Robots often lack full awareness of environments with novel objects, leading to failures in global planners that account for positional uncertainties (\eg, object location errors) but overlook manipulative space constraints (\eg, joint limits, end-effector accessibility). Rule-based methods like those in~\cite{jiao2022sequential} offer partial solutions via handcrafted heuristics but lack adaptability for long-horizon reasoning. As shown in \cref{fig:exception}, we introduce a local planner module to handle four motion planning exceptions. This taxonomy addresses two key limitations: 1)~While global planning considers positional uncertainty, it neglects constraints from robotic embodiment (\eg, reachability, stability). Our classification systematically captures these interaction constraints; 2)~The exception types align with standard motion planner diagnostics, allowing direct mapping to native planner outputs (\eg, \ac{vkc}'s collision checking~\cite{jiao2021efficient})—a critical feature for real-time exception handling in manipulation tasks. The prompts for guiding the \ac{llm} in inserting corrective actions for these exceptions during task execution are summarized in \cref{fig:prompts}.

\begin{figure}[t!]
    \centering 
    \framedtext{
        \small
        \textrm{\small 
            \vspace{-1mm} \\
            \textbf{System message}: As an assistive robot, you must implement corrective actions to address errors that arise during task executions. \\
            \textbf{Task description:} You will resolve errors during task executions using the following primitives: \hllb{\texttt{Pick}(x, y) to pick item x from y, \texttt{Place}(x, y) to place item x on y, \texttt{Open}(x) to open x, and \texttt{Close}(x) to close x. Additionally, you may temporarily place objects in a designated parking area.}\\
            \textbf{Example:} \textit{The motion exception is: "Placing $v_1$ on $v_2$ fails because $v_1$ collides with $v_3$. The parking place is $v_0$."\\
            \hllo{\textbf{Analysis:} Object $v_1$ collides with $v_3$, so $v_3$ must be adjusted to allow $v_1$ to be placed on $v_2$.\\
            \textbf{Robot Hand State:} Since the failed action was a placement, my hand is occupied with $v_1$. \\
            \textbf{Steps to Resolve:}\\
                1. Place $v_1$ in the parking area: \texttt{Place}(1, 0).\\
                2. Remove the collision by picking $v_3$: \texttt{Pick}(3, 2).\\
                3. Move $v_3$ to the parking area: \texttt{Place}(3, 0).\\
                4. Pick $v_1$ from the parking area: \texttt{Pick}(1, 0).\\
                5. Retry the failed action by placing $v_1$ on $v_2$: \texttt{Place}(1, 2).
            } \\
            I summarize the action sequence: [\texttt{Place}(1, 0), \texttt{Pick}(3, 2), \texttt{Place}(3, 0), \texttt{Pick}(1, 0), \texttt{Place}(1, 2)]}\\
            \textbf{Question:} You attempted to $\langle failure\_action \rangle$, but it failed due to $\langle exception \rangle$. The parking place is $\langle parking\_place\rangle$.
        }
    }
    \caption{\textbf{Prompt templates used by LLMPlanner.} \hllo{\textit{Chain of thought (CoT)}} prompts, and \hllb{\textit{action primitives}} prompts.} 
    \label{fig:prompts}
    \vspace{-16pt}
\end{figure}

\setstretch{0.975}

\section{Simulation and Experiment} \label{sec:sim}

In simulations, we conduct an ablation study of the \epog framework on five complex, long-horizon daily object transportation tasks across realistic scenes from the ProcThor-10k dataset~\cite{deitke2022}. Then, we illustrate the planned action sequences, highlighting \epog's efficiency in task execution. We finally validate the \epog framework through experiments on a physical robotic mobile manipulator with perception, confirming its effectiveness in real-world scenarios.

\subsection{Simulation Setup}

We evaluate five complex, long-horizon daily object transportation tasks, summarized in \cref{tab:task_description}: 1) Breakfast Preparation, 2) Bedroom Work, 3) Movie and Snack Preparation, 4) Tea Making and Relaxation, and 5) Bath Preparation. A total of 46 scenes were filtered from the ProcThor-10k dataset, each containing task-relevant objects. Four types of exceptions were randomly generated near these objects in each scene, with two exceptions present per scene. In the study, each task is performed once for each scene, and the following metrics are evaluated: \textbf{\%SR:} The percentage of successfully completed tasks out of the total scenes. \textbf{\%EN:} The average percentage difference in newly explored nodes. \textbf{\%TD:} The average percentage difference in travel distance (in meters) required to complete the task in each scene. Both \%EN and \%TD are calculated relative to the results of \epog.

\begin{table}[!ht]
\centering
\caption{Benchmark tasks and their goal predicates.}
\label{tab:task_description}
\footnotesize
\renewcommand{\arraystretch}{1.3} 
\setlength{\tabcolsep}{5pt}       

\newcommand{\g}[2]{\texttt{\{#1\}} $\to$ \texttt{#2}}
\newcommand{\gs}[2]{\texttt{#1} $\to$ \texttt{#2}}

\begin{tabularx}{\columnwidth}{@{} c >{\raggedright\arraybackslash}l c >{\raggedright\arraybackslash}X @{}}
\toprule
\textbf{No.} & \textbf{Task} & \textbf{Scenes} & \textbf{Goal Conditions} \\
\midrule
1 & Breakfast Prep. & 10 & 
\g{apple, bread, fork}{plate} \newline 
\gs{plate}{diningtable} \\

2 & Bedroom Work & 10 & 
\g{alarmclock, CD, laptop, pencil}{desk} \\

3 & Movie \& Snack & 10 & 
\gs{remote}{sofa} \newline 
\gs{bread}{plate}; \gs{plate}{diningtable} \\

4 & Tea \& Relax & 10 & 
\gs{kettle}{countertop}; \gs{cup}{diningtable} \newline 
\gs{remote}{sofa} \\

5 & Bath Prep. & 6 & 
\g{soapbottle, cloth}{faucet} \\
\bottomrule
\end{tabularx}
\end{table}

\begin{table*}[t!]
\centering
\caption{Ablation Study}
\label{tab:ablation}
\renewcommand\arraystretch{1.2}
\resizebox{\textwidth}{!}{%
\begin{tabular}{ccccccccccccccccccc}
\toprule
\multirow{2}{*}{\textbf{Method}} &
  \multicolumn{3}{c}{\textbf{Breakfast Prep.}} &
  \multicolumn{3}{c}{\textbf{Bedroom Work}} &
  \multicolumn{3}{c}{\textbf{Movie and Snack Prep.}} &
  \multicolumn{3}{c}{\textbf{Tea Making \& Relax.}} &
  \multicolumn{3}{c}{\textbf{Bath Prep.}} &
  \multicolumn{3}{c}{\textbf{Total}} \\ \cmidrule(lr){2-4}\cmidrule(lr){5-7}\cmidrule(lr){8-10}
            \cmidrule(lr){11-13}\cmidrule(lr){14-16}\cmidrule(lr){17-19}
 &
  \textbf{\%SR}$\uparrow$ & \textbf{\%EN}$\downarrow$ &  \textbf{\%TD}$\downarrow$ &  \textbf{\%SR}$\uparrow$ &  \textbf{\%EN}$\downarrow$ &  \textbf{\%TD}$\downarrow$ &  \textbf{\%SR}$\uparrow$ &  \textbf{\%EN}$\downarrow$ &  \textbf{\%TD}$\downarrow$ &  \textbf{\%SR}$\uparrow$ &  \textbf{\%EN}$\downarrow$ &  \textbf{\%TD}$\downarrow$ &  \textbf{\%SR}$\uparrow$ &  \textbf{\%EN}$\downarrow$ &  \textbf{\%TD}$\downarrow$ &  \textbf{\%SR}$\uparrow$ &  \textbf{\%EN}$\downarrow$ &  \textbf{\%TD}$\downarrow$ \\ \hline
\ac{llm} &
  10.0 &
  +0.0 &
  +110 &
  10.0 &
  +0.0 &
  +48.7 &
  10.0 &
  +0.0 &
  +179 &
  40.0 &
  +10.3 &
  +115 &
  16.7 &
  +0.0 &
  +16.7 &
  17.4 &
  +2.05 &
  +93.9 \\
Exp. + \ac{llm} &
  40.0 &
  +77.9 &
  +80.6 &
  40.0 &
  +75.9 &
  +84.0 &
  40.0 &
  +50.4 &
  +248 &
  60.0 &
  +75.0 &
  +124 &
  16.7 &
  +168 &
  +63.2 &
  41.3 &
  +89.4 &
  +120 \\
Exp. + PoG &
  100 &
  +77.2 &
  +59.0 &
  100 &
  +50.4 &
  +25.0 &
  \textbf{100} &
  +66.5 &
  +112 &
  \textbf{100} &
  +67.9 &
  +122 &
  \textbf{100} &
  +104 &
  +60.5 &
  \textbf{100} &
  +73.2 &
  +75.6 \\ \hline
\textbf{\epog (ours)} &
  \textbf{100} &
  \textbf{52.1} &
  \textbf{35.8} &
  \textbf{100} &
  \textbf{50.8} &
  \textbf{52.1} &
  80.0 &
  \textbf{52.9} &
  \textbf{33.0} &
  90.0 &
  \textbf{52.1} &
  \textbf{23.5} &
  83.3 &
  \textbf{46.4} &
  \textbf{43.6} &
  91.3 &
  \textbf{51.9} &
  \textbf{37.8} \\ \bottomrule
\end{tabular}%
}
\end{table*}

\subsection{Ablation Study}
To evaluate embodied task planning under environmental uncertainty systematically, we designed three baselines for the ablation study:

\begin{itemize}[leftmargin=*, noitemsep, nolistsep]
    \item The \textbf{\ac{llm} Planner} isolates data-driven planning by testing how \ac{llm}s overcome geometric infeasibility via iterative self-correction without explicit environment modeling—a scenario common in zero-shot deployments~\cite{song2023llm,yao2022react}.
    \item The \textbf{Exploration+\ac{llm} Planner} variant proactively explores the environment with optimized travel distance, establishing an upper bound for perception-planning pipelines in the absence of environmental priors. This configuration evaluates the extent to which exploration mitigates manipulative uncertainty.
    \item The \textbf{Exploration+PoG} variant implements a traditional rule-based planner with explicit exception handling from~\cite{jiao2022sequential}, serving as a reference for planners that rely on predefined spatiotemporal constraints and revealing differences in handling open-world task variance.
\end{itemize}

Our baseline design shows that planners integrating \ac{llm}s, including the \textbf{Exploration+PoG}, leverage the \ac{llm}'s open-world understanding to facilitate execution efficiency. However, \ac{llm}s may suffer from hallucinations and reasoning inconsistencies in complex or uncertain environments, whereas traditional symbol-based planners tend to achieve higher success rates due to their deterministic, rule-following nature.

\cref{tab:ablation} summarizes the results. Overall, the \ac{llm} planner exhibits lower success rates, particularly in long-horizon planning, as it often fails (with a maximum of 20 retries allowed per scene) to predict the next action correctly, such as walking to the wrong location for pick or place, or failing to find all task-relevant objects. This is because the \ac{llm} struggles to reason over complex, information-rich scene graphs and use them effectively for planning. Additionally, even in successful cases, the \ac{llm} planner generates less efficient plans for higher \%TD. In contrast, \epog achieves an overall success rate of 91.3\%, with a reduction of 40.0\% in explored nodes and 36.2\% in travel distance compared to the Exploration+PoG baseline. This is due to \epog's integrated manipulation planning and informed exploration strategy, which results in lower \%EN and \%TD. \epog continuously updates the belief graph with \ac{llm} heuristics and observations, optimizing both exploration and task execution. However, \epog still encounters 8.7\% planning failures under compound exception conditions. Although \epog's global planning algorithm guarantees complete task accomplishment through graph editing operations, its local planning module manifests cognitive ambiguities when handling compound exceptions induced by combinatorial object configurations. Despite minor performance losses in the local planner, the \ac{llm}-based local planner effectively handles random exceptions without requiring manually defined rules.

\subsection{Case Study}

\begin{figure*}[t!]
\centering
\includegraphics[width=\linewidth]{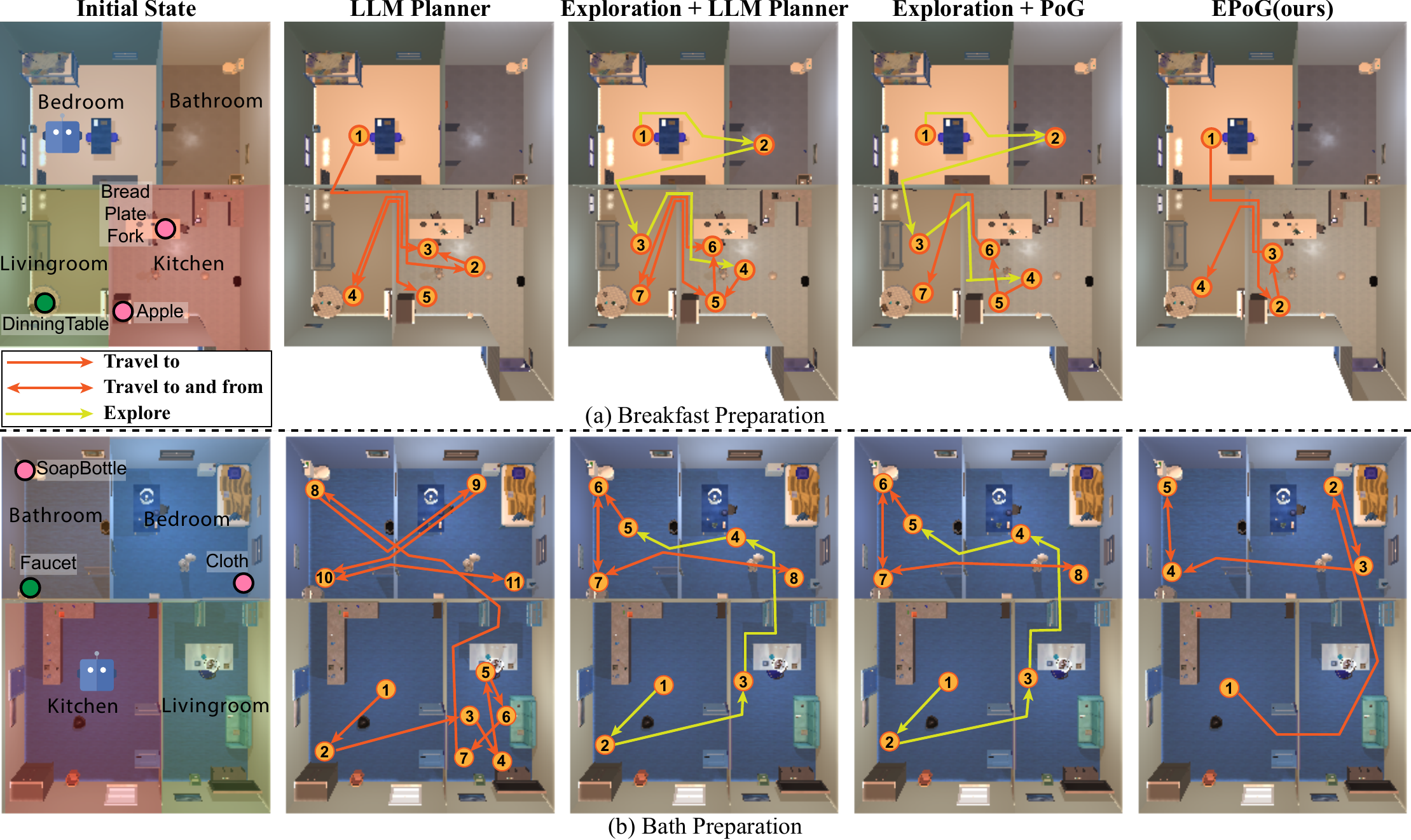}
\caption{Illustrations of the robot’s paths for two tasks, highlighting key locations and actions.}
\label{fig:case_study}
\vspace{-16pt}
\end{figure*}

\cref{fig:case_study} provides a graphical illustration of the robot's paths during two tasks. For clarity, only movements between \texttt{Rooms} or \texttt{Receptacles} are shown, omitting actions performed on \texttt{Objects} within the same \texttt{Receptacle}. In \cref{fig:case_study}(a), \epog demonstrated efficient task execution by first placing items on a plate in the kitchen ($\circled{1}\to\circled{2}\to\circled{3}$), then transporting the plate to the living room ($\to\circled{4}$). In contrast, the \ac{llm}-based method did not effectively use the plate as a transport tool, instead opting to carry objects separately, leading to a longer action sequence and increased travel distance. In \cref{fig:case_study}(b), \epog initially used the \ac{llm}'s commonsense knowledge to incorrectly estimate the \texttt{cloth}'s location on the nightstand. However, \epog enabled the robot to update the belief graph dynamically with new observations and regenerate a task plan. The robot then efficiently located the \texttt{cloth} in a cabinet and completed the task. Both tasks demonstrate \epog’s ability to seamlessly integrate exploration and planning, resulting in efficient task execution by reducing overall travel distance. In contrast, exploration-first strategies led the robot to explore unnecessary locations. The \ac{llm} planner struggled with the complexity of large scene graphs, resulting in unnecessary travel between irrelevant locations and increased total travel distance. Although the PoG method optimized the plan, the separation of exploration and planning led to higher execution effort due to longer travel distances.

\subsection{Experiment Setup}
We conducted two real-world experiments on a physical mobile manipulator to evaluate \epog's ability to plan with unknown object states and resolve exceptions during task execution. Notably, our physical robot experiments focus on validating the task planning performance of the \epog framework in real-world environments, while deliberately abstracting away implementation details of robot perception. The experiments utilized \ac{vkc}-based mobile manipulation planning~\cite{jiao2021efficient,li2024dynamic} and Curobo~\cite{sundaralingam2023curobo} to generate real-time whole-body trajectories for object interactions.

\begin{figure}[th!]
\centering
\includegraphics[width=\linewidth]{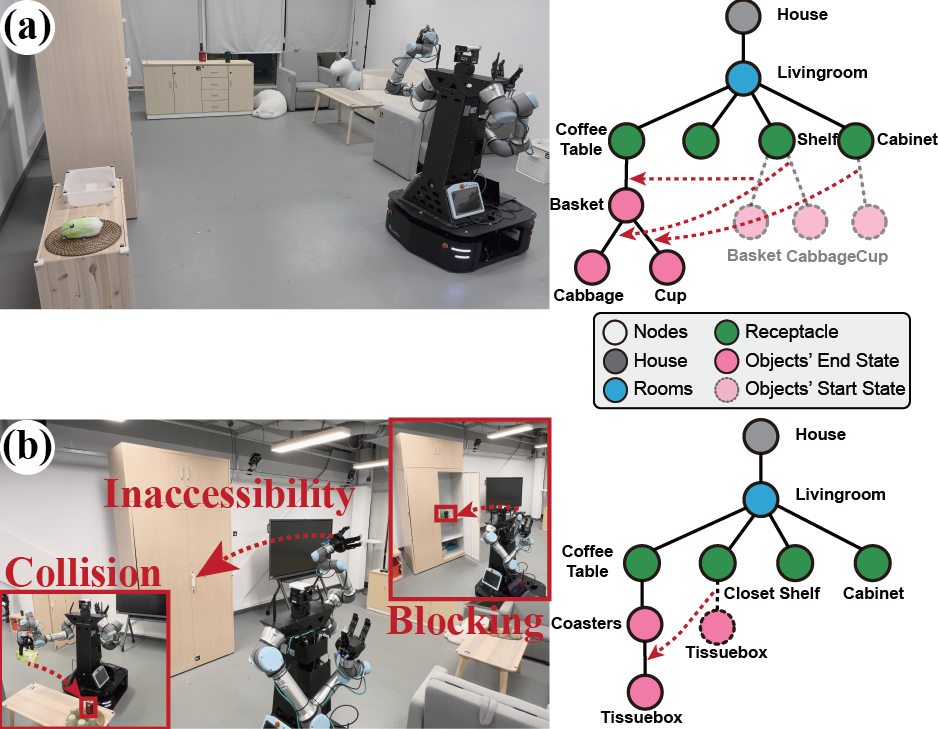}
\caption{\textbf{Real-world experiments.} (a) The robot must place three objects (with one unknown object) on a coffee table. (b) The robot must retrieve a tissue box from a closet and place it on a coaster, first removing a cup and a tea can that obstruct task execution. Object states in the closet and on the coffee table are initially unknown.}
\label{fig:exp_setup}
\vspace{-16pt}
\end{figure}

\cref{fig:exp_setup} illustrates the task setup. In the first experiment (\cref{fig:exp_setup}(a)), the robot is tasked with moving a basket onto a coffee table and placing a toy cabbage and a cup into the basket. Initially, the robot is aware of the cabbage and basket states, but the cup's location is unknown. Using \ac{llm}, the robot infers that the cup might be on the coffee table, and the global planner of \epog generates an optimized task plan to minimize travel distance. The robot first places the cabbage in the basket, moves the basket to the table, and then plans to retrieve the cup from the coffee table and place it into the basket. However, when the cup is not found on the table, the robot updates its belief graph and ultimately locates the cup on top of a cabinet, successfully completing the task.
In the second experiment (\cref{fig:exp_setup}(b)), the robot is instructed to place a tissue box on a coaster on the coffee table. The robot must handle exceptions like inaccessibility, blocking, and potential collisions during the task. The \ac{llm} correctly estimates the tissue box to be inside a closed closet, but the contents of the closet are unknown. After opening the closet, the robot discovers that a cup is blocking the tissue box, so it temporarily moves the cup to a shelf. Similarly, when a tea can obstructs the coaster, the robot sets down the tissue box, removes the tea can, and successfully places the box on the coaster. Throughout the process, \epog primarily relies on its local planning module to manage these exceptions effectively. The robot's execution details are documented in the supplementary video.

\section{Conclusion} \label{sec:conclusion}
In conclusion, we presented \epog, a planning framework that effectively integrates exploration and task planning within unknown environments using graph-based representations. By utilizing a bi-level planning scheme, \epog combines graph-based sequential planning with \ac{llm}-based local replanning, enabling robust long-horizon operation and effective handling of unexpected situations. Studies across daily complex object transportation tasks demonstrated \epog’s efficiency in task execution and potential for real-world robotic applications. Future integration of a 3D scene graph perception module, along with the incorporation of \ac{tamp} methods~\cite{jiao2021consolidated,jiao2025integration}, could further improve its practicality, advancing robots’ capabilities in real-world scenarios.

\bibliographystyle{ieeetr}
\bibliography{IEEEfull}

\end{document}